\documentclass[runningheads]{llncs}
\usepackage{orcidlink}
\usepackage{adjustbox}
\usepackage{subcaption}

\usepackage{graphicx}%
\usepackage{multirow}%
\usepackage{amsmath,amssymb,amsfonts}%

\usepackage{mathrsfs}%
\usepackage[title]{appendix}%
\usepackage{xcolor}%
\usepackage{textcomp}%
\usepackage{manyfoot}%
\usepackage{booktabs}%
\usepackage{algorithm}%
\usepackage{algorithmicx}%
\usepackage{algpseudocode}%
\usepackage{listings}%

\usepackage[T1]{fontenc}
\usepackage{graphicx}
\usepackage{orcidlink}

\begin{document}
\title{FACROC: a fairness measure for FAir Clustering through ROC curves}
%
%
\author{Tai Le Quy\inst{1}\orcidlink{0000-0001-8512-5854} \and
Long Le Thanh \inst{2} \orcidlink{0009-0007-8971-0648}\and
Lan Luong Thi Hong \inst{3}\orcidlink{0000-0002-4083-2253}  \and
Frank Hopfgartner\inst{1,4}\orcidlink{0000-0003-0380-6088} }
\authorrunning{Le Quy et al.}
%
\institute{University of Koblenz, Germany \\
\email{tailequy@uni-koblenz.de}\\
\and
Hanoi University of Science and Technology, Vietnam\\
\email{attentionocr@gmail.com}
\and
Hanoi University of Industry, Vietnam\\
\email{lanlhbk@haui.edu.vn}
\and
University of Sheffield, United Kingdom \\
\email{hopfgartner@uni-koblenz.de}
}
\maketitle              
\begin{abstract}
Fair clustering has attracted remarkable attention from the research community. Many fairness measures for clustering have been proposed; however, they do not take into account the clustering quality w.r.t. the values of the protected attribute. In this paper, we introduce a new visual-based fairness measure for fair clustering through ROC curves, namely FACROC. This fairness measure employs AUCC as a measure of clustering quality and then computes the difference in the corresponding ROC curves for each value of the protected attribute. Experimental results on several popular datasets for fairness-aware machine learning and well-known (fair) clustering models show that FACROC is a beneficial method for visually evaluating the fairness of clustering models.

\end{abstract}
\section{Introduction}
\label{sec:introduction}
Clustering is a fundamental problem in unsupervised learning, and fairness in clustering has garnered significant attention within the machine learning (ML) community, starting with the foundational work of Chierichetti et al. \cite{chierichetti2017fair}. Fair clustering techniques aim to ensure equitable representation or treatment of groups or individuals within clusters. Researchers have focused on two main challenging problems in fair clustering: defining and enforcing fairness constraints \cite{chhabra2021overview}. Hence, a number of fairness notions and techniques were introduced to ensure fairness constraints in clustering and can be applied in many domains, such as healthcare \cite{chakrabarti2022new}, education \cite{le2023review}, etc.

There are more than 20 fairness notions in fair clustering \cite{chhabra2021overview}. Since fairness notions can be turned into measures, we will use the terms ``fairness notion'' and ``fairness measure'' interchangeably. Fairness notions are defined based on the group-level, individual-level, algorithm agnostic, and algorithm specific. At the group-level, the algorithms should not discriminate against or unfairly favor any group of individuals in the predictions. For instance, \textit{balance} notion \cite{chierichetti2017fair}, the most popular fairness notion used for fair clustering, requires a ratio balance between the protected group, e.g., female, and the non-protected group, e.g., male. Unlike group-level fairness, individual-level fairness ensures that similar individuals are treated similarly by the clustering model. The fairness notion of \textit{proportionality} for centroid clustering \cite{chen2019proportionally} is an example of individual-level fairness. However, to our knowledge, all the defined fairness notions do not consider the clustering quality w.r.t. values of the protected attribute.

Regarding the clustering quality, it is determined by several measures, such as the silhouette coefficient, sum of squared error (SSE), Dunn index (DI), and others \cite{le2023review,saxena2017review}. However, most of these metrics are difficult to visualize for comparison between clustering models. To address this issue, AUCC (Area Under the Curve for Clustering) \cite{jaskowiak2022area} has been introduced as a measure of clustering quality. This is a visual-based measure that utilizes the Receiver Operating Characteristics (ROC) analysis of clustering results. In addition, ABROCA, a visual-based fairness measure, has been proposed for the classification problem \cite{gardner2019evaluating}. This method evaluates the fairness of classification models through slicing analysis based on the ROC curves. ABROCA measures the absolute value of the area between the ROC curves of the protected and non-protected groups.

To this end, we propose a new fairness measure for FAir Clustering through ROC curves (shortly: FACROC) which takes into account the clustering quality on each value of the protected attribute. In particular, we use AUCC to measure the clustering quality and then compute the FACROC by the deviation between the ROC curves corresponding to each value of the protected attribute. Afterward, we perform the experiments on five popular datasets used in fairness-aware ML with three prevalent fair clustering models to evaluate the performance of FACROC versus other popular fairness measures.

The rest of our paper is structured as follows: Section \ref{sec:relatedwork} overviews the related work. The computation of the FACROC measure is described in Section \ref{sec:facroc}. Section \ref{sec:evaluation} presents the details of our experiments on various datasets and clustering models. Finally, the conclusion and outlook are summarized in Section \ref{sec:conclusion}.

\section{Related work}
\label{sec:relatedwork}
In this work, we focus on two main types of fairness measures for clustering including group-level fairness and individual-level fairness. Regarding group-level, \emph{balance} is the first group fairness measure introduced by \cite{chierichetti2017fair}. Next, the \emph{bounded representation} measure was proposed with the aim of reducing imbalances in cluster representations of protected attributes \cite{ahmadian2019clustering}. This measure was generalized with two parameters $\alpha$ and $\beta$ in the study of \cite{bera2019fair}. Subsequently, \emph{social fairness} notion was introduced by \cite{ghadiri2021socially}, which aims to provide equitable costs for different clusters. Additionally, \emph{diversity-aware} fairness was initiated by \cite{thejaswi2021diversity} which ensures a minimum number of cluster centers in the clustering are selected from each group. Based on the summarization task, the \emph{fair summaries} measure was used to ensure that the data summary for each group is represented equally \cite{kleindessner2019fair}.

 In contrast to group-level fairness, individual-level fairness notions aim to ensure that similar individuals are treated similarly. For example, Chen et al. \cite{chen2019proportionally} proposed the individual-level fairness notion of proportionality for centroid clustering to guarantee that points are treated equally. An individual-level concept that establishes a fair radius for clusters in center-based clustering objectives was presented by Jung et al. \cite{jung2020service}. Chakrabarti et al. \cite{chakrabarti2022new} provided algorithms for the $k$-center objective and proposed the idea of individual fairness, which guarantees that points receive comparable quality of service.
 
However, unlike other fairness measures for clustering based on the representation of protected attribute values in clusters, we propose a new visual-based fairness measure that takes into account the difference in the clustering performance for each group w.r.t. the protected attribute.

\section{FACROC: a fairness measure for fair clustering} \label{sec:facroc}
Given a dataset $\mathcal{X} = \{x_1,\ldots,x_n\}$ with $n$ data points and a clustering $\mathcal{C}=\{C_1, C_2,..., C_k\}$ with $k$ clusters, the corresponding AUCC value is computed with the following steps \cite{jaskowiak2022area}:
\begin{enumerate}
    \item Compute a similarity matrix of the objects in the original dataset.
    \item Acquire two arrays that present the pairwise relationship for each pair of objects:
        \begin{enumerate}
            \item Similarity: get from the similarity matrix.
            \item Clustering: 1 if the pair belongs to the same cluster; otherwise, 0.
        \end{enumerate}
    \item AUCC is obtained from the ROC analysis procedure with two above arrays, in which pairwise clustering memberships correspond to the ``true classes'' concept in classification.
\end{enumerate}

Given a binary protected attribute $P = \{p, \Bar{p}\}$; e.g., gender=\{female, male\}, we inherit the concept of fairness from the ABROCA fairness measure for classification \cite{gardner2019evaluating}: ``\emph{equal model performance across subgroups}''. To this end, FACROC is defined as the absolute value of the area between the protected ($ROC_{p}$) and non-protected group ($ROC_{\overline{p}}$) curves across all possible thresholds $t \in [0,1]$ of False Positive Rate (FPR) and True Positive Rate (TPR). The absolute difference between the two curves is measured to capture the case that the curves may cross each other.
\begin{equation}
\label{eq:facroc}
    \int_{0}^{1}\mid ROC_{p}(t) - ROC_{\overline{p}}(t)\mid \,dt.
\end{equation}

\begin{figure}
    \centering
    \includegraphics[width=0.5\linewidth]{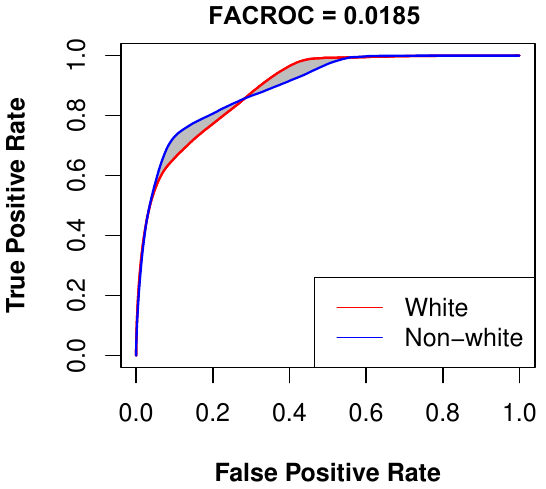}
    \caption{An example of the FACROC slice plot}
    \label{fig:FACROC-sample}
    \vspace{-10pt}
\end{figure}

The value range: $FACROC \in [0, 1]$. The lower value indicates a lower difference in the clustering quality between the two groups and, therefore, a fairer model. Fig. \ref{fig:FACROC-sample} visualizes the FACROC slice plot on the COMPAS dataset.

\section{Evaluation}
\label{sec:evaluation}
In this section, we evaluate the performance of (fair) clustering models on well-known fairness measures for clustering and our proposed FACROC fairness measure on prevalent fairness-aware datasets.
\vspace{-5pt}
\subsection{Datasets}
\label{subsec:dataset}
We perform the experiments on five popular datasets for fairness-aware ML \cite{lequy2022survey}, as summarized in Table \ref{tbl:dataset}\footnote{Abbreviations: F (Female), M (Male), W (White), NW (Non-White)}.
\vspace{-15pt}
\begin{table*}[!h]
\centering
\caption{An overview of the datasets}\label{tbl:dataset}
\begin{adjustbox}{width=1\textwidth}
\begin{tabular}{lccccc}

\hline
\multicolumn{1}{c}{\textbf{ Datasets }} &  
\multicolumn{1}{c}{\textbf{ \#Instances}} &
\multicolumn{1}{c}{\begin{tabular}[c]{@{}c@{}}\textbf{ \#Instances}\\\textbf{  (cleaned)} \end{tabular}} &
\multicolumn{1}{c}{\textbf{ \#Attributes }} &
\multicolumn{1}{c}{\begin{tabular}[c]{@{}c@{}}\textbf{Protected attribute}\end{tabular}} & 
\multicolumn{1}{c}{\textbf{ k* }}\\ \hline
Adult  & 48,842 & 45,222 & 15 & Gender (F: 14,695; M: 30,527) & 2 \\ 
COMPAS & 4,743 & 4,020 & 51  & Race (NW: 2,561; W: 1,459) & 7 \\
Credit card  & 30,000 & 30,000 & 24  & Gender (F: 18,112; M: 11,888) & 2 \\
German credit  & 1000 & 1000 & 21 & Gender (F: 310; M:690) &  2 \\
Student-Mat.  & 395 & 395 & 33 & Gender (F: 208, M: 187 ) & 9  \\
Student-Por.  & 649 & 649 & 33 & Gender (F: 383; M: 266) & 9 \\
\hline
\end{tabular}
\end{adjustbox}
\vspace{-5pt}
\end{table*}

In particular, the \textit{\textbf{Adult dataset}}\footnote{\url{https://archive.ics.uci.edu/dataset/2/adult}} is one of the most prevalent datasets for fairness-aware ML research. The class attribute (whether the income is greater than 50,000\$) is removed because this study uses the dataset for clustering tasks.
The \textit{\textbf{COMPAS dataset}}\footnote{\url{https://github.com/propublica/compas-analysis}} is used for crime recidivism risk prediction. We convert the protected attribute \emph{Race} into a binary attribute with values \emph{\{White, Non-White\}}. We remove datetime attributes (\emph{compas\_screening\_date, dob}, etc.), defendants' name/ID, and the class label \emph{two years recidivism}. 
The \textit{\textbf{Credit card clients dataset}}\footnote{\url{https://archive.ics.uci.edu/ml/datasets/default+of+credit+card+clients}} contains information about 30,000 customers in Taiwan in October 2005. The prediction task is to forecast whether a customer will default in the next month. We remove the class label for clustering.
The \textit{\textbf{German credit dataset}}\footnote{\url{https://archive.ics.uci.edu/dataset/144/statlog+german+credit+data}} contains information about 1000 customers, with the goal of predicting whether a customer has good or bad credit. We also eliminate this class label in our experiments.
The \textit{\textbf{Student performance dataset}}\footnote{\url{https://archive.ics.uci.edu/dataset/320/student+performance}} details students' academic performance in secondary education at two Portuguese schools in the 2005–2006 school year, covering two subsets: Mathematics (shortly: Student-Mat.) and Portuguese (shortly: Student-Por.).

\subsection{Experimental setups}
\label{subsec:setup}
\textbf{Clustering models}.
We evaluate the performance of traditional $k$-means and hierarchical clustering, as well as well-known fair clustering models.
\begin{itemize}
\item \textbf{Fair clustering through fairlets} \cite{chierichetti2017fair} (shortly: Fairlet): This is the first work on fair clustering at the group-level to ensure an equal representation of each value of the protected attribute in each cluster. A two-phase approach was proposed: 1) Fairlet decomposition: grouping all instances into ``fairlets'' which are small clusters that satisfy the fairness constraint; 2) Clustering on fairlets: applying standard clustering methods, such as $k$-center, $k$-median, to these fairlets to produce the final fair clusters. 

\item \textbf{Scalable fair clustering} \cite{backurs2019scalable} (shortly: Scalable): This is an extended investigation of the fair $k$-median clustering problem \cite{chierichetti2017fair}, with a new practical approximate fairlet decomposition algorithm that runs in nearly linear time. Therefore, this proposed approach can be applied to large datasets.

\item \textbf{Proportionally fair clustering} \cite{chen2019proportionally} (shortly: Proportionally):  The authors define proportional fairness: any group of $n/k$ points should have the right to form their cluster if there exists a center closer for all $n/k$ points. The goal is to find clustering where no subset of points has a justified complaint about their assigned cluster without assuming predefined protected groups.
\end{itemize}

\noindent\textbf{Fairness measures}.
We compare the proposed FACROC measure with the following well-known fairness measures:
\begin{itemize}
\item \textbf{Balance} \cite{chierichetti2017fair}:
Given a clustering $\mathcal{C}=\{C_1, C_2,..., C_k\}$ with $k$ clusters, $\mathcal{X}$ be a set of data points, $P = \{p, \Bar{p}\}$ be the protected attribute, $\psi : \mathcal{X} \rightarrow P$ denotes the demographic group to which the data point belongs, i.e., male or female, the balance of clustering $\mathcal{C}$ is computed by:
\begin{equation}
\label{eq:measure_balance}
    balance(\mathcal{C}) = min_{i=1}^{k} balance(C_i).
\end{equation}
where: 
\begin{equation}
\label{eq:cluster_balance}
    balance(C_i) = \min\left(\frac{|\{x \in C_i \mid \psi(x) = p\}|}{|\{x \in C_i \mid \psi(x) = \Bar{p}\}|}, \frac{|\{x \in C_i \mid \psi(x) = \Bar{p}\}|}{|\{x \in C_i \mid \psi(x) = p\}|}\right).
\end{equation}


\item \textbf{Proportionality} \cite{chen2019proportionally}: Given $\mathcal{X}$ be a set of $n$ data points, $\mathcal{Y}$ be a set of feasible cluster centers $m$ and a number $\rho > 1$, we call a clustering $\mathcal{C} \subseteq \mathcal{Y}$ ($|\mathcal{C}| = k$) is $\rho$-\emph{proportional} if $\forall S \subseteq \mathcal{X}$ with $|S| \geq \lceil \frac{n}{k} \rceil$ and for all $y \in \mathcal{Y}$, there exists $i \in S$ with $\rho \cdot d(i, y) \geq D_i(\mathcal{C})$, where $d(i,y)$ is the distance between points $i$ and $y$, and $D_i(\mathcal{C}) = min_{x\in \mathcal{C}}d(i, x)$. When $\rho = 1$, we call this proportional fairness. 

\end{itemize}

\begin{figure*}
    \centering
    \includegraphics[width=1.0\textwidth]{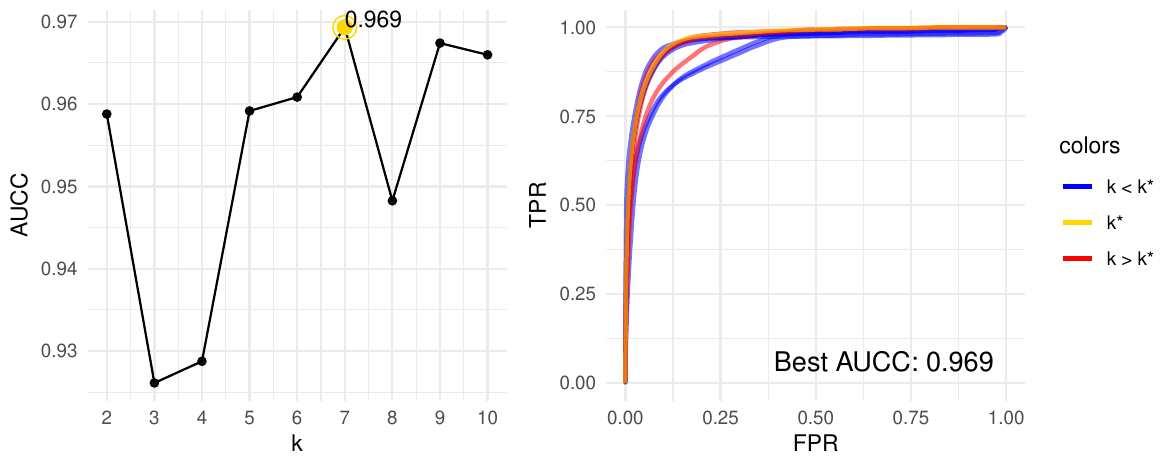}
    \caption{COMPAS: selecting the optimal number of cluster ($k$*) by AUCC}
    \label{fig:compas-k}
    \vspace{-10pt}
\end{figure*}

\noindent\textbf{Parameter selection}.
We denote the optimal number of clusters as $k^*$, which is determined based on the AUCC values using the implementation provided by~\cite{jaskowiak2022area}. Fig. \ref{fig:compas-k} illustrates the value of $k^*$ for the COMPAS dataset; other values of $k^*$ are presented in Table \ref{tbl:dataset}. The minimum balance threshold is set to 0.4 because the minimum balance score of all datasets is 0.4493 (corresponding to the German credit dataset).

\subsection{Experimental results}
\label{subsec:results}
We present the results of clustering models regarding performance (Silhouette coefficient and AUCC) and fairness (Balance, Proportionality, and FACROC). The source code is available at \url{https://github.com/tailequy/FACROC}.

\begin{figure*}[!h]
\centering
\begin{subfigure}{.32\linewidth}
    \centering
    \includegraphics[width=\linewidth]{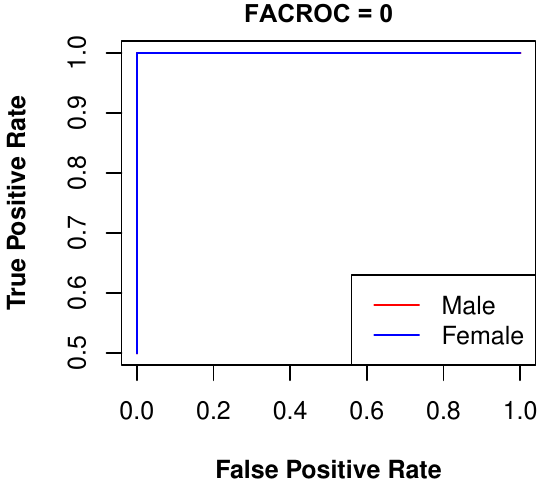}
    \caption{$k$-means}
\end{subfigure}
\hfill
\begin{subfigure}{.32\linewidth}
    \centering
    \includegraphics[width=\linewidth]{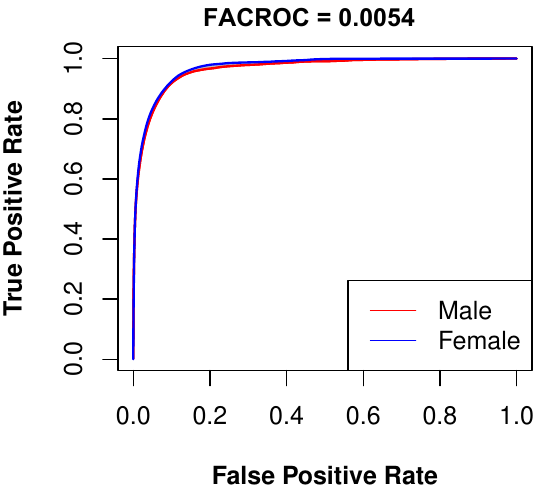}
    \caption{Hierarchical}
\end{subfigure}    
\hfill
\begin{subfigure}{.32\linewidth}
    \centering
    \includegraphics[width=\linewidth]{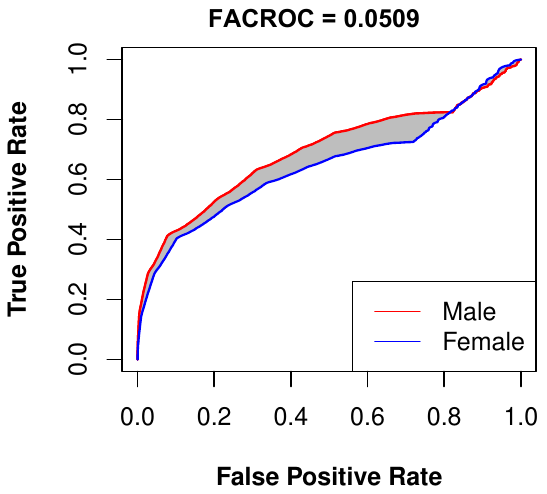}
    \caption{Fairlet}
\end{subfigure}    
\bigskip
\begin{subfigure}{.32\linewidth}
    \centering
    \includegraphics[width=\linewidth]{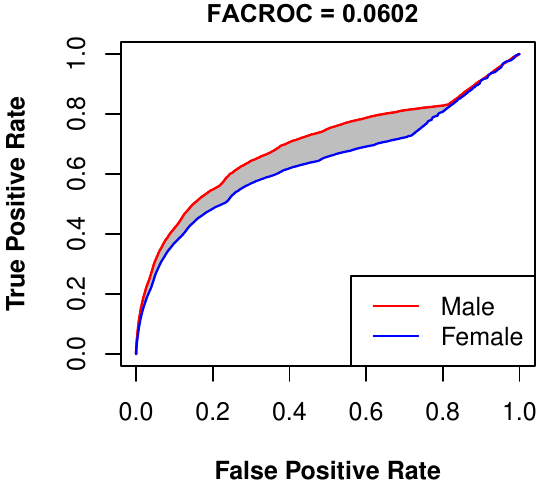}
    \caption{Scalable}
\end{subfigure}
\begin{subfigure}{.32\linewidth}
    \centering
    \includegraphics[width=\linewidth]{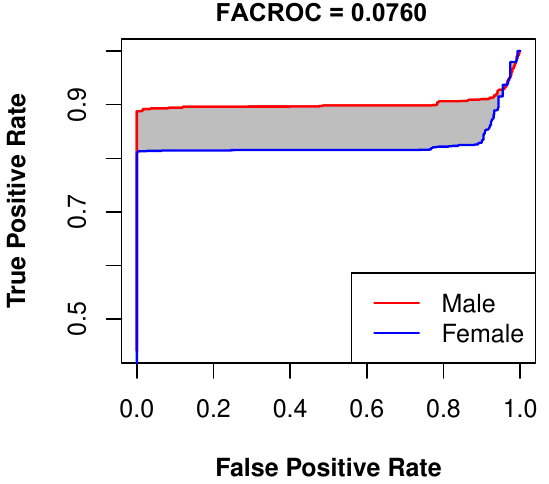}
    \caption{Proportionally}
\end{subfigure}
\caption{Adult: FACROC slice plots}
\label{fig:adult_facroc}
\end{figure*}

\textbf{Adult dataset.}
The performance of the clustering models is shown in Table \ref{tbl:adult} and Fig. \ref{fig:adult_facroc}, with the best results highlighted in \textbf{bold}. In terms of clustering performance, obviously, $k$-means outperforms fair clustering methods. However, regarding fairness constraint, fair clustering models outperform $k$-means w.r.t. the fairness measure that they optimize. In detail, \emph{Fairlet} and \emph{Scalable} have better balance scores, while the \emph{Proportionally} model is better in terms of the proportionality measure. This is explained by the fact that the definition of fairness is different in all models. Interestingly, the FACROC value of $k$-means is perfect, while the \emph{Proportionally} model shows the worst result, i.e., the performances of observed fair clustering models are biased toward groups of people.
\begin{table}[!h]
\centering
\caption{Adult: performance of (fair) clustering models}\label{tbl:adult}
\begin{tabular}{lccccc}
\hline
\textbf{Measures} &  
\multicolumn{1}{c}{\textbf{$k$-means}} & 
\multicolumn{1}{c}{\textbf{Hierarchical}} & 
\multicolumn{1}{c}{\textbf{Fairlet}} &
\multicolumn{1}{c}{\textbf{Scalable}} & 
\multicolumn{1}{c}{\textbf{Proportionally}} \\ \hline
Silhouette coefficient       & \textbf{0.9861} & \textbf{0.9861} & 0.4062 & 0.4377 & 0.3711\\
AUCC                   & \textbf{1.0000} & 0.9998 & 0.6607 & 0.6569 & 0.8503 \\
Balance                & 0.1684 & 0.1926 & \textbf{0.5001} & 0.4396 & 0.2966\\
Proportionally         & 1.0000 & 1.3274 & 1.4701 & 1.5994 & \textbf{1.6321}\\
FACROC                 & \textbf{0.0000} & 0.0054 & 0.0509 & 0.0602 & 0.0760 \\
\hline
\end{tabular}
\vspace{-5pt}
\end{table}

\begin{figure*}[!h]
\centering
\vspace{-5pt}
\begin{subfigure}{.32\linewidth}
    \centering
    \includegraphics[width=\linewidth]{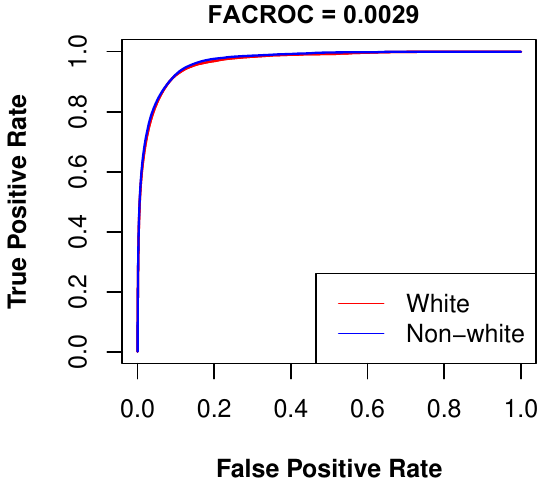}
    \caption{$k$-means}
\end{subfigure}
\hfill
\begin{subfigure}{.32\linewidth}
    \centering
    \includegraphics[width=\linewidth]{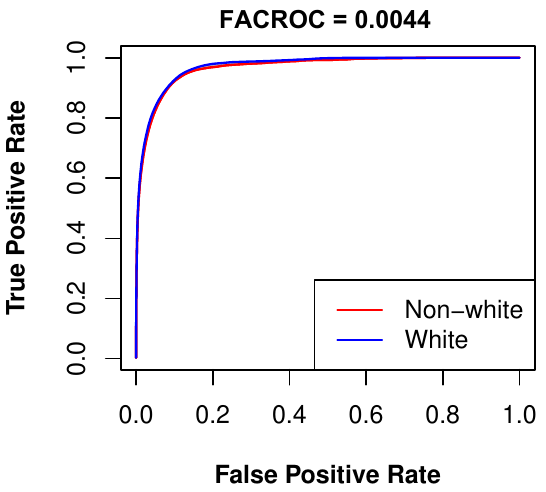}
    \caption{Hierarchical}
\end{subfigure}    
\hfill
\begin{subfigure}{.32\linewidth}
    \centering
    \includegraphics[width=\linewidth]{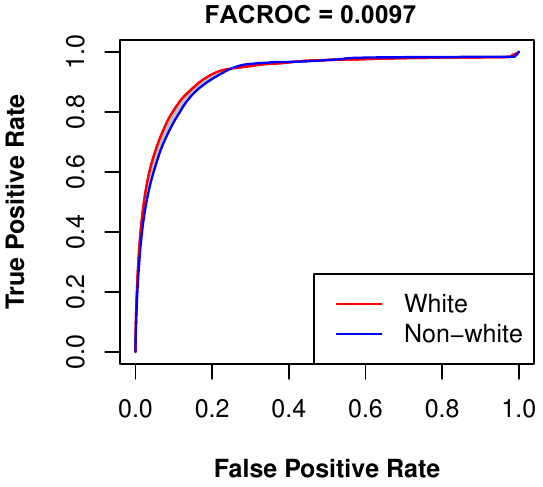}
    \caption{Fairlet}
\end{subfigure}    
\hfill
\begin{subfigure}{.32\linewidth}
    \centering
    \includegraphics[width=\linewidth]{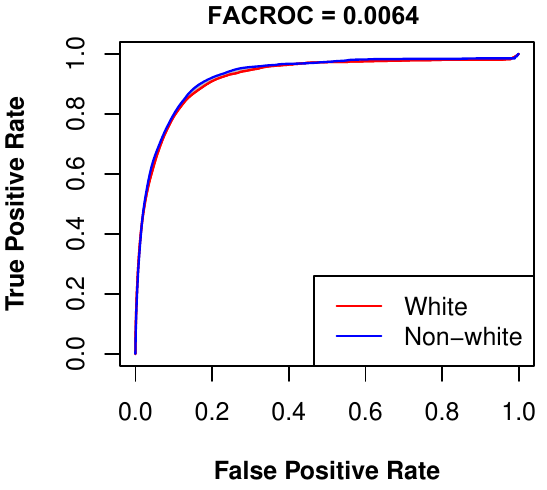}
    \caption{Scalable}
\end{subfigure}
\begin{subfigure}{.32\linewidth}
    \centering
    \includegraphics[width=\linewidth]{prop.compas.facroc.pdf}
    \caption{Proportionally}
\end{subfigure}
\caption{COMPAS: FACROC slice plots}
\label{fig:compas_facroc}
\end{figure*}

\textbf{COMPAS dataset}.
An interesting trend is also observed in the results of the clustering models in the COMPAS data set (Table \ref{tbl:compas} and Fig. \ref{fig:compas_facroc}) when the FACROC of the $k$-means algorithm is significantly better than that of other models. This can be attributed to fair clustering models prioritizing fairness constraints, which can result in variations in clustering quality among groups.
\begin{table}[!h]
\centering
\caption{COMPAS: performance of (fair) clustering models}\label{tbl:compas}
\begin{tabular}{lccccc}
\hline
\textbf{Measures} &  
\multicolumn{1}{c}{\textbf{$k$-means}} & 
\multicolumn{1}{c}{\textbf{Hierarchical}} & 
\multicolumn{1}{c}{\textbf{Fairlet}} &
\multicolumn{1}{c}{\textbf{Scalable}} & 
\multicolumn{1}{c}{\textbf{Proportionally}}\\ \hline
Silhouette coefficient& \textbf{0.6110} & 0.6082 & 0.3827 & 0.3583 & 0.3903 \\
AUCC                   & \textbf{0.9690} & 0.9689 & 0.9233 & 0.9236 & 0.8933 \\
Balance                & 0.0000 & 0.1017 & 0.4186 & \textbf{0.4291} & 0.3226 \\
Proportionality        & 0.8554 & 1.1728 & 0.9443 & 1.0937 & \textbf{1.2861} \\
FACROC                 & \textbf{0.0029} & 0.0044 & 0.0097 & 0.0064 & 0.0185 \\
\hline
\end{tabular}
\vspace{-5pt}
\end{table}

\begin{figure*}[!h]
\centering
\begin{subfigure}{.32\linewidth}
    \centering
    \includegraphics[width=\linewidth]{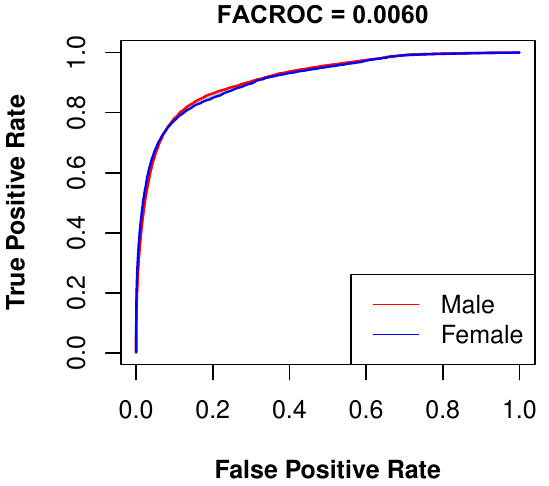}
    \caption{$k$-means}
\end{subfigure}
\hfill
\begin{subfigure}{.32\linewidth}
    \centering
    \includegraphics[width=\linewidth]{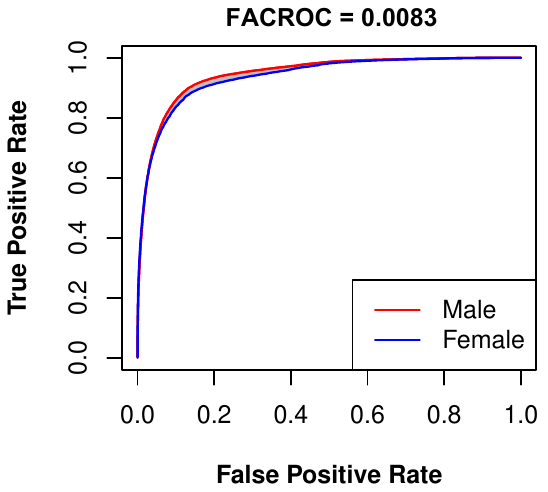}
    \caption{Hierarchical}
\end{subfigure}    
\hfill
\begin{subfigure}{.32\linewidth}
    \centering
    \includegraphics[width=\linewidth]{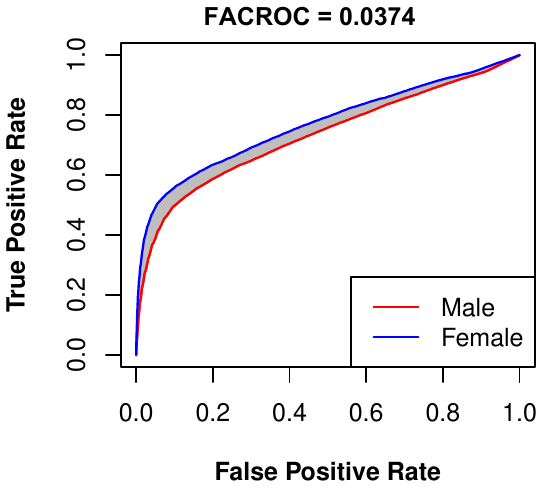}
    \caption{Fairlet}
\end{subfigure}    
\hfill
\begin{subfigure}{.32\linewidth}
    \centering
    \includegraphics[width=\linewidth]{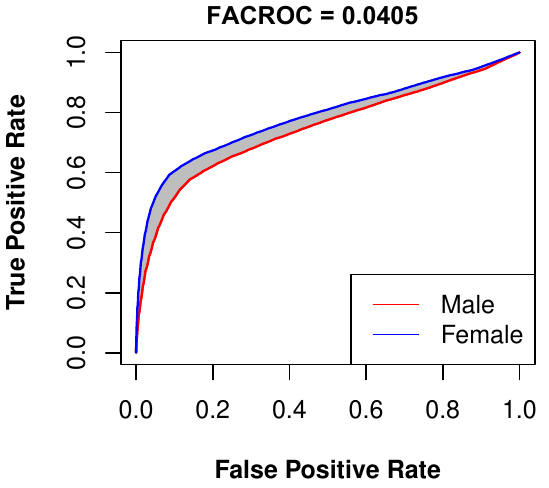}
    \caption{Scalable}
\end{subfigure}
\begin{subfigure}{.32\linewidth}
    \centering
    \includegraphics[width=\linewidth]{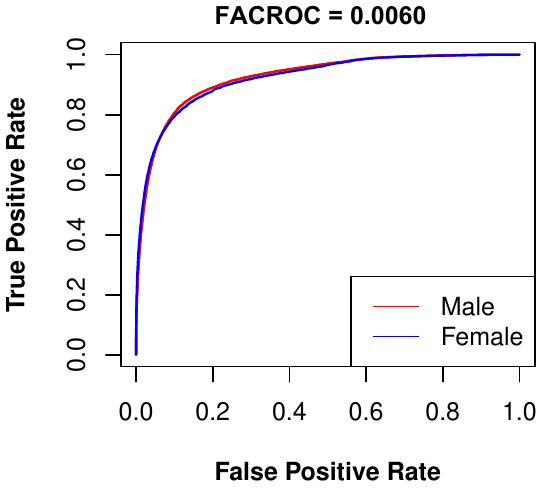}
    \caption{Proportionally}
\end{subfigure}
\caption{Credit card: FACROC slice plots}
\label{fig:creditcard_facroc}
\end{figure*}

\textbf{Credit card clients dataset}. Compared with fair clustering models, $k$-means is still the model with the highest clustering quality (Table \ref{tbl:creditcard} and Fig. \ref{fig:creditcard_facroc}). However, in terms of the FACROC measure, \emph{Propotionally} and $k$-means share the top rank with a very low value.

\begin{table}[!h]
\centering
\caption{Credit card clients: performance of (fair) clustering models}\label{tbl:creditcard}
\begin{tabular}{lccccc}
\hline
\textbf{Measures} &  
\multicolumn{1}{c}{\textbf{$k$-means}} & 
\multicolumn{1}{c}{\textbf{Hierarchical}} & 
\multicolumn{1}{c}{\textbf{Fairlet}} &
\multicolumn{1}{c}{\textbf{Scalable}} & 
\multicolumn{1}{c}{\textbf{Proportionally}}\\ \hline
Silhouette coefficient & 0.5778 & \textbf{0.6014} & 0.3390 & 0.3476 & 0.5528 \\
AUCC                   & 0.9247 & \textbf{0.9422} & 0.7456 & 0.7607 & 0.9127 \\
Balance                & 0.6145 & 0.5927 & \textbf{0.6957} & 0.6440 & 0.6477 \\
Proportionality        & 1.0003 & 0.9939 & 1.0909 & 1.0828 & \textbf{1.4935} \\
FACROC                 & \textbf{0.0060} & 0.0083 & 0.0374 & 0.0405 & \textbf{0.0060} \\
\hline
\end{tabular}
\end{table}

\begin{figure*}[!h]
\centering
\begin{subfigure}{.32\linewidth}
    \centering
    \includegraphics[width=\linewidth]{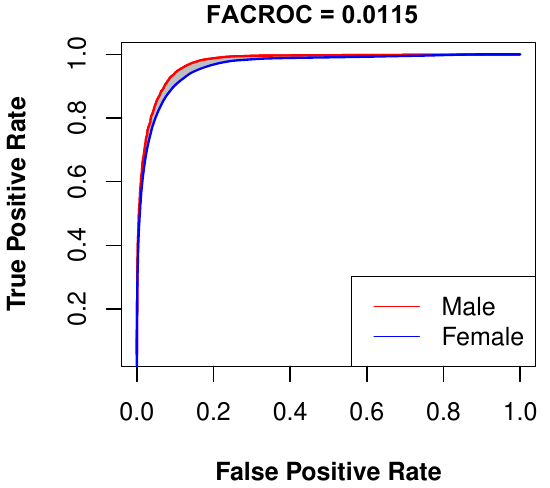}
    \caption{$k$-means}
\end{subfigure}
\hfill
\begin{subfigure}{.32\linewidth}
    \centering
    \includegraphics[width=\linewidth]{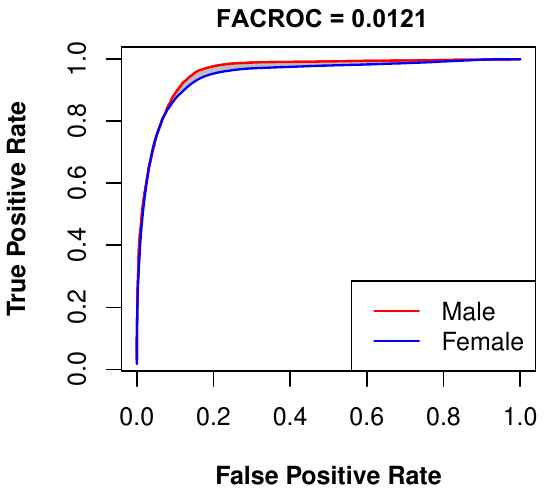}
    \caption{Hierarchical}
\end{subfigure}    
\hfill
\begin{subfigure}{.32\linewidth}
    \centering
    \includegraphics[width=\linewidth]{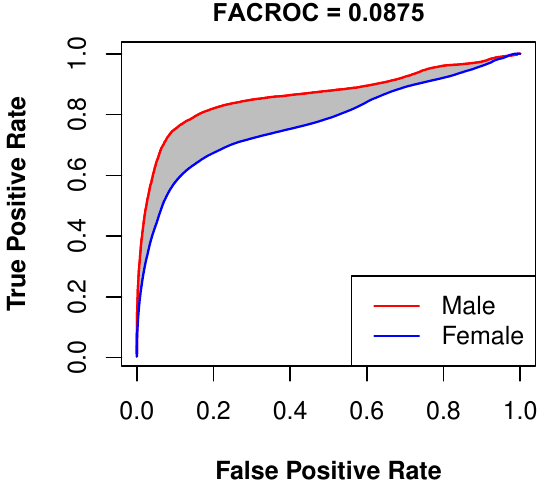}
    \caption{Fairlet}
\end{subfigure}    
\hfill
\begin{subfigure}{.32\linewidth}
    \centering
    \includegraphics[width=\linewidth]{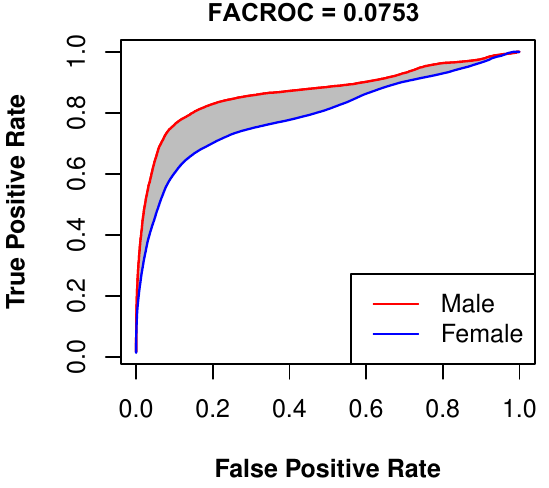}
    \caption{Scalable}
\end{subfigure}
\begin{subfigure}{.32\linewidth}
    \centering
    \includegraphics[width=\linewidth]{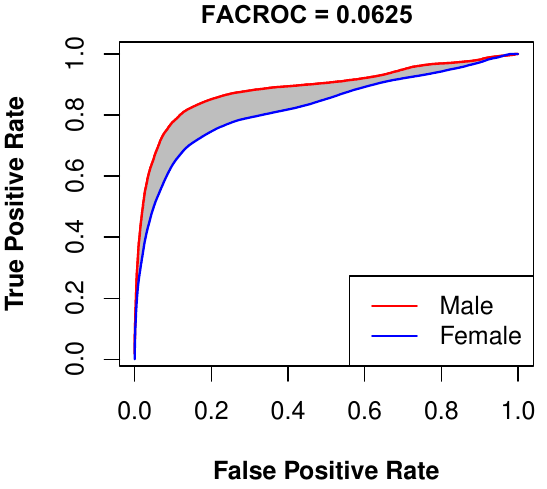}
    \caption{Proportionally}
\end{subfigure}
\caption{German credit: FACROC slice plots}
\label{fig:german_facroc}
\end{figure*}

\textbf{German credit}.
In this dataset, a similar trend is observed in Table \ref{tbl:german-credit} and Fig. \ref{fig:german_facroc}. $k$-means achieves the best results according to the FACROC measure, while fair clustering models have better results on the fairness measure they are designed to optimize.
\begin{table}[!h]
\centering
\caption{German credit: performance of (fair) clustering models}\label{tbl:german-credit}
\begin{tabular}{lccccc}
\hline
\textbf{Measures} &  
\multicolumn{1}{c}{\textbf{$k$-means}} & 
\multicolumn{1}{c}{\textbf{Hierarchical}} & 
\multicolumn{1}{c}{\textbf{Fairlet}} &
\multicolumn{1}{c}{\textbf{Scalable}} & 
\multicolumn{1}{c}{\textbf{Proportionally}}\\ \hline
Silhouette coefficient & \textbf{0.7222} & 0.6963 & 0.1387 & 0.3053 & 0.5302 \\
AUCC                   & \textbf{0.9672} & 0.9523 & 0.8046 & 0.8192 & 0.8440 \\
Balance                & 0.3008 & 0.3314 & \textbf{0.4545} & 0.4045 & 0.3669 \\
Proportionality        & 0.9469 & 0.9874 & 1.2964 & 1.2975 & \textbf{1.3829} \\
FACROC                 & \textbf{0.0115} & 0.0121 & 0.0875 & 0.0753 & 0.0625 \\
\hline
\end{tabular}
\vspace{-8pt}
\end{table}

\begin{figure*}[!h]
\centering
\begin{subfigure}{.32\linewidth}
    \centering
    \includegraphics[width=\linewidth]{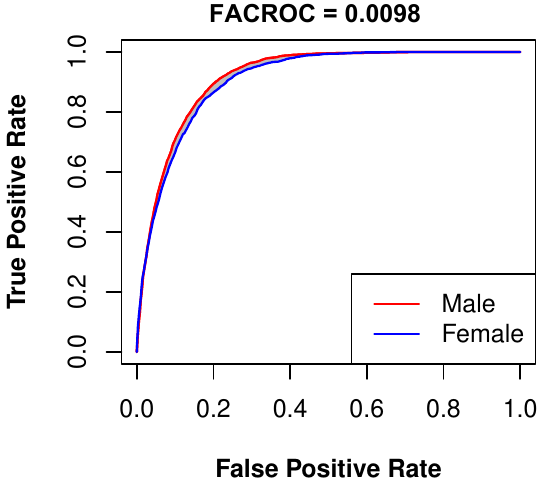}
    \caption{$k$-means}
\end{subfigure}
\hfill
\begin{subfigure}{.32\linewidth}
    \centering
    \includegraphics[width=\linewidth]{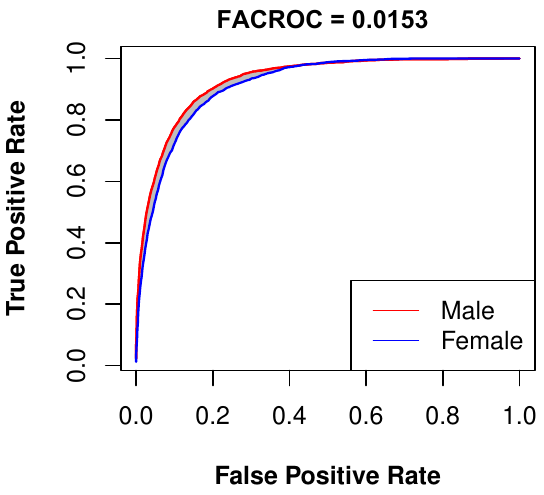}
    \caption{Hierarchical}
\end{subfigure}    
\hfill
\begin{subfigure}{.32\linewidth}
    \centering
    \includegraphics[width=\linewidth]{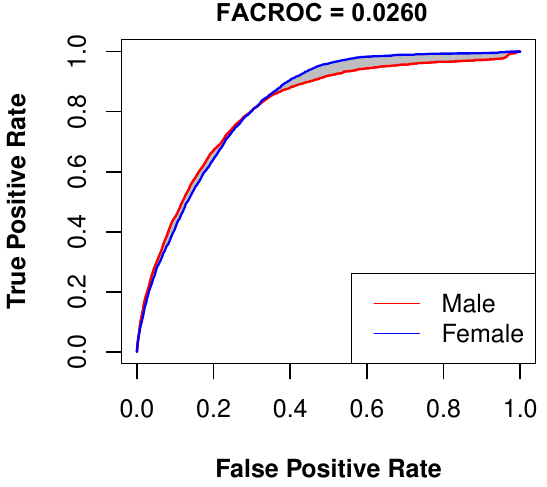}
    \caption{Fairlet}
\end{subfigure}    
\begin{subfigure}{.32\linewidth}
    \centering
    \includegraphics[width=\linewidth]{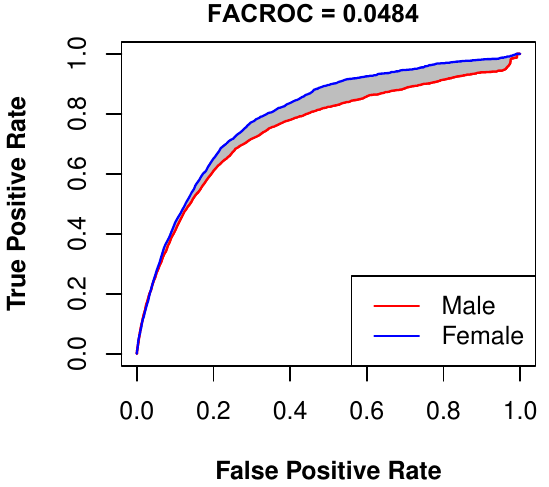}
    \caption{Scalable}
\end{subfigure}
\begin{subfigure}{.32\linewidth}
    \centering
    \includegraphics[width=\linewidth]{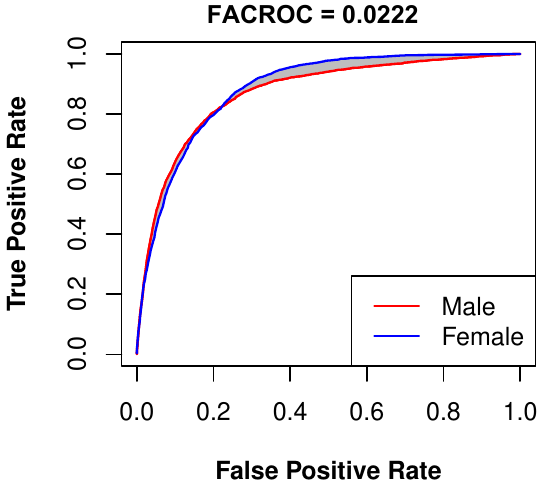}
    \caption{Proportionally}
\end{subfigure}
\caption{Student-Mat: FACROC slice plots}
\label{fig:student_mat_facroc}
\end{figure*}

\begin{figure*}[!h]
\centering
\begin{subfigure}{.32\linewidth}
    \centering
    \includegraphics[width=\linewidth]{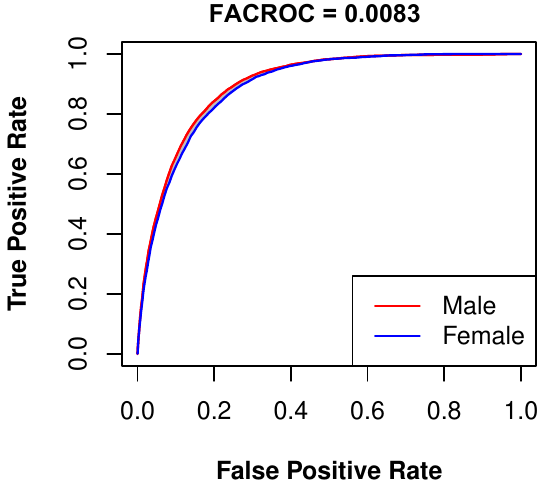}
    \caption{$k$-means}
\end{subfigure}
\hfill
\begin{subfigure}{.32\linewidth}
    \centering
    \includegraphics[width=\linewidth]{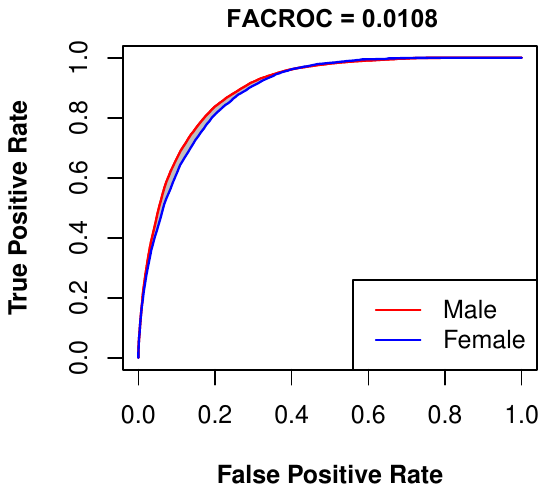}
    \caption{Hierarchical}
\end{subfigure}    
\hfill
\begin{subfigure}{.32\linewidth}
    \centering
    \includegraphics[width=\linewidth]{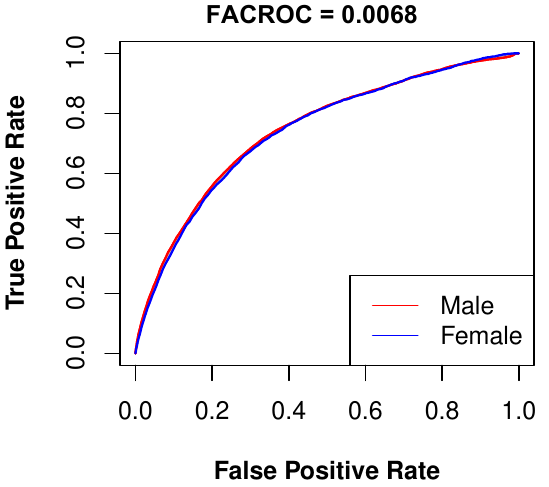}
    \caption{Fairlet}
\end{subfigure}    
\hfill
\begin{subfigure}{.32\linewidth}
    \centering
    \includegraphics[width=\linewidth]{scalable.studentmat.facroc.pdf}
    \caption{Scalable}
\end{subfigure}
\begin{subfigure}{.32\linewidth}
    \centering
    \includegraphics[width=\linewidth]{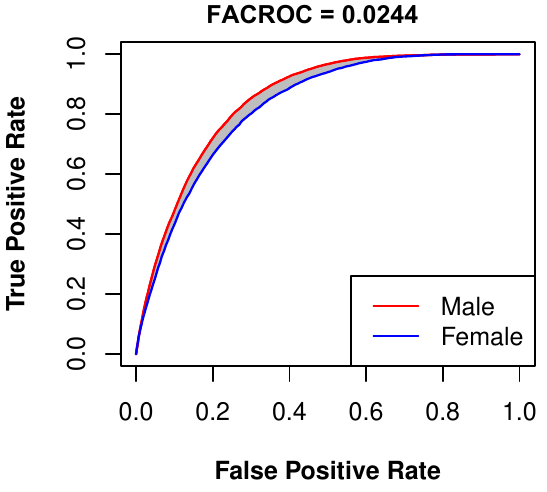}
    \caption{Proportionally}
\end{subfigure}
\caption{Student-Por: FACROC slice plots}
\label{fig:student_por_facroc}
\end{figure*}

\textbf{Student performance dataset}.
In the Student-Mat subset (Table \ref{tbl:student_mat} and Fig. \ref{fig:student_mat_facroc}), in terms of clustering performance, 
$k$-means outperforms other clustering models, although its silhouette coefficient is low. Interestingly, \emph{Scalable} fair clustering achieves the highest balance score due to having the lowest AUCC score. Moreover, the 
$k$-means model once again achieves the highest FACROC value, followed by \emph{Hierarchical clustering} and \emph{Proportionally} fair clustering. However, a different trend emerges in the Student-Por subset (Table \ref{tbl:student_por} and Fig. \ref{fig:student_por_facroc}), where \emph{Fairlet} outperforms other models in terms of FACROC, while \emph{Proportionally} achieves the best balance score and proportionality.
\begin{table}[!h]
\centering
\caption{Student-Mat: performance of (fair) clustering models}\label{tbl:student_mat}
\begin{tabular}{lccccc}
\hline
\textbf{Measures} &  
\multicolumn{1}{c}{\textbf{$k$-means}} & 
\multicolumn{1}{c}{\textbf{Hierarchical}} & 
\multicolumn{1}{c}{\textbf{Fairlet}} &
\multicolumn{1}{c}{\textbf{Scalable}} & 
\multicolumn{1}{c}{\textbf{Proportionally}} \\ \hline
Silhouette coefficient & \textbf{0.1814} & 0.1526 & 0.0456 & 0.0552 & 0.1111 \\
AUCC                   & \textbf{0.9117} & 0.8931 & 0.8201 & 0.7705 & 0.8823 \\
Balance                & 0.0000 & 0.1294 & 0.4231 & \textbf{0.6176} & 0.3182 \\
Proportionality        & 0.9970 & 1.1295 & 1.2782 & 1.1510 & \textbf{1.4167} \\
FACROC                 & \textbf{0.0098} & 0.0153 & 0.0260 & 0.0484 & 0.0222 \\
\hline
\end{tabular}
\vspace{-8pt}
\end{table}

\begin{table}[!h]
\centering
\caption{Student-Por: performance of (fair) clustering models}\label{tbl:student_por}
\begin{tabular}{lccccc}
\hline
\textbf{Measures} &  
\multicolumn{1}{c}{\textbf{$k$-means}} & 
\multicolumn{1}{c}{\textbf{Hierarchical}} & 
\multicolumn{1}{c}{\textbf{Fairlet}} &
\multicolumn{1}{c}{\textbf{Scalable}} & 
\multicolumn{1}{c}{\textbf{Proportionally}} 
\\ \hline
Silhouette coefficient & \textbf{0.1345} & 0.1136 & 0.0242 & 0.0591 & 0.0618 \\
AUCC                   & \textbf{0.8935} & 0.8641 & 0.7389 & 0.8270 & 0.8459 \\
Balance                & 0.3881 & 0.3972 & 0.4184 & 0.4231 & \textbf{0.4483} \\
Proportionality        & 1.0341 & 1.3728 & 1.2142 & 1.1623 & \textbf{1.4320} \\
FACROC                 & 0.0083 & 0.0108 & \textbf{0.0068} & 0.0484 & 0.0244 \\
\hline
\end{tabular}
\end{table}

\section{Conclusions and outlook}
\label{sec:conclusion}
In this work, we introduced FACROC, a new fairness notion for fair clustering that leverages the ROC curves of clustering analysis w.r.t. a protected attribute. We evaluated our proposed fairness measure on several datasets and (fair) clustering models. The results demonstrate that our measure is effective in visualizing and analyzing the fairness of clustering models using ROC curves. Furthermore, the evaluation highlights significant variations among fairness measures due to differences in their definitions and objective functions. In the future, we plan to extend this work by developing a method to optimize the FACROC value and adapt it for multiple protected attributes.

%
%
%
 \bibliographystyle{splncs04}
 \bibliography{bibliography}
\end{document}